\documentclass[lettersize,journal]{IEEEtran}
\usepackage{amsmath,amsfonts}
\usepackage{algorithmic}
\usepackage{algorithm}
\usepackage{array}
\usepackage[caption=false,font=normalsize,labelfont=sf,textfont=sf]{subfig}
\usepackage{textcomp}
\usepackage{stfloats}
\usepackage{url}
\usepackage{verbatim}
\usepackage{graphicx}
\usepackage{multirow}
\usepackage{booktabs}
\usepackage{xcolor}
\usepackage{cite}
\begin{document}

\title{TinySplat: Feedforward Approach for Generating Compact 3D Scene Representation}

\author{
	Zetian~Song,
        Jiaye~Fu,
	Jiaqi~Zhang,
        Xiaohan~Lu,\\
        Chuanmin~Jia, ~\IEEEmembership{Member,~IEEE,}
	Siwei~Ma, ~\IEEEmembership{Fellow,~IEEE,}
	Wen~Gao, ~\IEEEmembership{Fellow,~IEEE,}\\	
		\thanks{
		}
        \thanks{Zetian~Song, Jiaqi~Zhang, Xiaohan~Lu, Siwei~Ma and Wen Gao are with the State Key Laboratory of Multimedia Information Processing, School of Computer Science, Peking University, Beijing, 100871, China~(e-mail: songzt@pku.edu.cn, jqzhang@pku.edu.cn, luxiaohan@stu.pku.edu.cn, swma@pku.edu.cn, wgao@pku.edu.cn). }
        \thanks{Jiaye~Fu is with the School of Electronic and Computer Engineering, Peking University, Shenzhen, 518055, China, and also with the State Key Laboratory of Multimedia Information Processing, School of Computer Science, Peking University, Beijing, 100871, China (email: jyfu@stu.pku.edu.cn).}
        \thanks{Chuanmin~Jia is with the Wangxuan Institute of Computer Technology, State Key Laboratory of Multimedia Information Processing, Peking University, Beijing 100871, China~(email: cmjia@pku.edu.cn).}
	}

\markboth{IEEE Transactions on Circuits and Systems for Video Technology}%
{Shell \MakeLowercase{\textit{et al.}}: A Sample Article Using IEEEtran.cls for IEEE Journals}


\maketitle

\begin{abstract}
The recent development of feedforward 3D Gaussian Splatting~(3DGS) presents a new paradigm to reconstruct 3D scenes. Using neural networks trained on large-scale multi-view datasets, it can directly infer 3DGS representations from sparse input views. Although the feedforward approach achieves high reconstruction speed, it still suffers from the substantial storage cost of 3D Gaussians. Existing 3DGS compression methods relying on scene-wise optimization are not applicable due to architectural incompatibilities.
To overcome this limitation, we propose TinySplat, a complete feedforward approach for generating compact 3D scene representations. Built upon standard feedforward 3DGS methods, TinySplat integrates a training-free compression framework that systematically eliminates key sources of redundancy. Specifically, we introduce View-Projection Transformation (VPT) to reduce geometric redundancy by projecting geometric parameters into a more compact space. We further present Visibility-Aware Basis Reduction (VABR), which mitigates perceptual redundancy by aligning feature energy along dominant viewing directions via basis transformation. Lastly, spatial redundancy is addressed through an off-the-shelf video codec.
Comprehensive experimental results on multiple benchmark datasets demonstrate that TinySplat achieves over 100$\times$ compression for 3D Gaussian data generated by feedforward methods. Compared to the state-of-the-art compression approach, we achieve comparable quality with only 6\% of the storage size. Meanwhile, our compression framework requires only 25\% of the encoding time and 1\% of the decoding time.


\end{abstract}


\section{Introduction}
\IEEEPARstart{T}{he}
3D Gaussian Splatting (3DGS) has emerged as a powerful technique for reconstructing 3D scenes from multi-view images. 
By explicitly representing a scene as a collection of anisotropic Gaussian primitives, 3DGS enables photorealistic rendering and supports real-time free-viewpoint navigation. 
The vanilla 3DGS formulation relies on stochastic gradient descent (SGD) and adaptive density control to jointly optimize the geometric and appearance attributes of the Gaussian primitives. 
While this approach achieves impressive visual fidelity, its reliance on per-scene optimization and dense input views significantly limits its scalability and practicality.

Recent feedforward 3DGS utilizes Neural Network~(NN) to directly infer the parameters of 3D Gaussian primitives from input images, thereby eliminating the need for iterative optimization. 
They offer substantial improvements in reconstruction speed and are capable of handling sparse view inputs, making them attractive for time-sensitive applications such as AR/VR and mobile 3D capture. 
However, the efficiency comes at the cost of significantly increased data volume. 
3DGS requires numerous Gaussian primitives to model a specific scene, resulting in substantial memory and storage demands.
This expansion poses major challenges for data transmission and interactive rendering, highlighting the need for effective compression techniques specialized for the unique characteristics of 3DGS data.

Prior research has explored optimization-based strategies for compressing 3DGS models, such as pruning~\cite{papantonakis2024reducing, lee2024compactrad, shi2024lapisgs, wang2024end2end, fan2024lightgaussian}, clustering~\cite{lu2024scaffold}, motion estimation~\cite{gao2024hicom, tang2025compressing}, and probabilistic modeling~\cite{chen2024hac, wang2024contextgs}. Notably, HAC~\cite{chen2024hac} employs multi-resolution hash grid priors for compact encoding, while ContextGS~\cite{wang2024contextgs} enhances compression efficiency through hierarchical context modeling. Nevertheless, due to their reliance on iterative optimization, these methods inherit the same limitations as vanilla 3DGS and are thus unsuitable for feedforward methods.

To overcome these limitations, FCGS~\cite{chen2024fcgc} introduced an optimization-free compression framework applicable to arbitrary 3D Gaussian models. By integrating hash grid-based hyperpriors and context priors through a multi-path entropy module, FCGS achieves notable compression ratios across diverse datasets. However, its effectiveness for feedforward-generated Gaussian models remains limited. First, FCGS does not fully exploit the inherent spatial redundancy derived from feedforward geometric generation. Additionally, the uniform treatment of feature channels hinders effective utilization of perceptual redundancy present in appearance features. Furthermore, reliance on multiple NN-based prior models introduces computational overhead unsuitable for real-time or edge-based scenarios.

In this paper, we propose \textbf{TinySplat}, a fully training-free approach for generating compact 3D Gaussian scene representations directly from multi-view images. TinySplat effectively addresses existing limitations by coupling a feedforward 3DGS generation stage with a rendering-aware compression stage. Specifically, in the Gaussian generation stage, we employ existing feedforward methods, such as DepthSplat~\cite{xu2024depthsplat}, to produce Gaussian feature maps, where each element defines parameters of individual Gaussian primitives. Subsequently, we introduce a novel compression framework designed to leverage spatial and perceptual redundancies within these feature maps for efficient storage and transmission.

In the compression framework, we propose \textit{View-Projection Transform} (\textbf{VPT}) to address structural redundancy from geometric generation. By exploiting the pixel-aligned characteristics of feedforward 3DGS, VPT reveals a more regular spatial layout, significantly enhancing local correlation and enabling more effective compression. Additionally, we propose the \textit{Visibility-Aware Basis Reduction} (\textbf{VABR}) method to handle perceptual redundancy in color attributes. Leveraging the anisotropic properties of spherical harmonic (SH) functions, VABR selectively retains perceptually dominant components while suppressing negligible ones, thereby improving compression efficiency without compromising rendering fidelity. Lastly, given the resemblance between Gaussian feature maps and traditional 2D images, we further exploit spatial redundancies using an off-the-shelf video codec.

Our main contributions are summarized as follows:
\begin{itemize}
    \item We introduce TinySplat, a fully feedforward pipeline consisting of Gaussian generation and a novel compression stage, enabling compact 3D Gaussian scene representations without scene-dependent optimization.

    \item We develop the VPT module, applying a coordinate-space transformation that exploits characteristics specific to feedforward geometric inference, thereby reducing structural redundancy and enhancing reconstruction quality.

    \item We propose the VABR module, which utilizes the visibility of SH bases to selectively preserve the most perceptually informative radiance components under constrained viewing directions, enabling highly compact color representations.

    \item Extensive experiments demonstrate that TinySplat can achieve comparable quality to DepthSplat with only 1\% of storage. In comparison to the state-of-the-art~(SoTA) compression method, TinySplat achieves a 90\% reduction in storage and a 75\% reduction in encoding time, simultaneously delivering superior visual fidelity..
\end{itemize}

The remainder of this paper is organized as follows. Section~\ref{sec:related works} provides an overview of related works on 3DGS generation and compression. Section~\ref{sec:method} describes the proposed TinySplat in detail. Section~\ref{sec:experiment} presents experimental evaluations and demonstrates the advantages of TinySplat. Section~\ref{sec:limitations} outlines limitations and future research directions. Finally, Section~\ref{sec:conclusion} concludes the paper.

\section{Related Works}
\label{sec:related works}
\subsection{Novel View Synthesis via Per-scene Optimization}
Novel-view synthesis is a pivotal task in computer vision and graphics, aiming to generate photorealistic views from captured multi-view images.
Recently, per-scene optimization methods have achieved remarkable progress by formulating the 3D representation as learnable parameters optimized using SGD.

As a seminal work in this field, the Neural Radiance Field~(NeRF)~\cite{mildenhall2021nerf} proposed by Mildenhall \textit{et al.} uses a deep multi-layer perceptron to learn color and opacity mappings. Novel view rendering is then performed through the volumetric rendering pipeline.
Muller \textit{et al.} further propose INGP~\cite{muller2022instant}, which models the scene using multi-resolution hash grids. By querying the explicit 3D structure, INGP can significantly reduce the computational cost of both training and inference.
Researchers have also explored various explicit data structures for scene representation, including voxel grids~\cite{sun2022direct, fridovich2022plenoxels}, multiple tensor planes~\cite{fridovich2023k, chen2022tensorf}, and octrees~\cite{yu2021plenoctrees}.
Based on NeRFs, Kerbl \textit{et al.} propose 3DGS~\cite{3dgs_vanilla}, which replaces volumetric rendering with a hardware-friendly rasterization technique. 3DGS enables real-time rendering on consumer devices and has inspired numerous follow-up works targeting different application domains~\cite{fu2024colmap, qian20243dgs, meng2024mirror, yu2024mip}. 


\subsection{Sparse View 3D Reconstruction}
Acquiring dozens of input views is usually impractical in real-world applications. As a result, researchers have investigated methods for 3D reconstruction from sparse input views. Several approaches~\cite{niemeyer2022regnerf, truong2023sparf, zhang2024cor, li2024dngaussian, han2024binocular} improve the 3DGS optimization process by introducing specialized regularization, such as view consistency and depth normalization. 
Meanwhile, some methods construct 3D scene representations using principles from multi-view stereo~\cite{pixelsplat, mvsnerf, chen2023explicit, szymanowicz2024flash3d, xu2024murf, mvsplat}. They utilize techniques like epipolar geometry and cost volume aggregation. Some methods also leverage Vision Transformer architectures to effectively fuse features across different views~\cite{noposplat}.
Further research also employs structural priors derived from large-scale diffusion models to maintain consistency between multiple viewpoints, enhancing the quality of novel view synthesis~\cite{gao2024cat3d, wu2024reconfusion, zhang2024gs, liu20243dgs, liu2024one, chen2024mvsplat360}. 

\subsection{feedforward 3D Gaussian Splatting}
Vanilla 3DGS~\cite{3dgs_vanilla} requires per-scene optimization, making it computationally intensive and time-consuming to obtain explicit scene representations. Recently, numerous feedforward 3DGS methods have been proposed, targeting 3D objects~\cite{szymanowicz2024splatter, tang2024lgm, zhou2024gpsplus, zheng2024gps, xu2024grm} or entire scenes~\cite{pixelsplat, noposplat, chen2024lara, szymanowicz2024flash3d, wewer2024latentsplat, min2024epipolar, chen2024mvsplat360, xu2024depthsplat, chen2024pref3r}. The feedforward methods focus on fast reconstruction of 3D scenes from sparse input views.
Specifically, these methods typically comprise a depth estimation module and a Gaussian attribute synthesis module. 

For single object reconstruction, GPS-Gaussian~\cite{zheng2024gps} and GPS-Gaussian+~\cite{zhou2024gpsplus} focus on 3D Gaussian reconstruction of the human body. Leveraging strong structural priors of human anatomy, these methods can generate a high-quality human body model from only a few input views. Szymanowicz \textit{et al.} propose Splatter Image~\cite{szymanowicz2024splatter}, which maps each input pixel to a 3D Gaussian using a simple yet efficient network, achieving real-time performance at 38 FPS for forward reconstruction. GRM~\cite{xu2024grm} further introduces a transformer-based architecture that effectively fuses multi-view information to achieve better reconstruction quality.

In contrast, full-scene reconstruction poses greater challenges than single-object reconstruction due to increased scene complexity and lack of reliable structural priors. To address these challenges, Charatan \textit{et al.} propose PixelSplat~\cite{pixelsplat}, which leverages deep learning to predict dense probability distributions in 3D space and samples Gaussian centers accordingly to enable fast scene reconstruction. Chen \textit{et al.} introduce MVSplat~\cite{mvsplat}, which generates multiple depth maps per view and computes cross-view confidence scores to achieve more accurate geometric reconstruction.
Zhang et al. propose the Gaussian Graph Network (GGN)~\cite{zhang2024gaussian} to improve reconstruction quality with abundant input views. By constructing a graph structure, GGN enables each Gaussian primitive to aggregate features from multiple viewpoints, which significantly reduces the number of Gaussian primitives and effectively suppresses artifacts. 
Xu \textit{et al.} further propose DepthSplat~\cite{xu2024depthsplat}, leveraging pre-trained monocular depth features to improve both depth prediction and reconstruction quality, which achieves SoTA performance.

\subsection{3D Gaussian Splatting compression}
The substantial computational cost of generating 3D Gaussian models highlights the need for efficient storage and transmission, making compression a key challenge in 3DGS research.
Researchers have proposed numerous training-based methods to compress 3D Gaussian structures and reduce their storage and transmission costs.

Some works create sparser representations by pruning less significant Gaussian primitives during training~\cite{papantonakis2024reducing, lee2024compactrad, shi2024lapisgs, wang2024end2end, fan2024lightgaussian}. 
Vector quantization is also widely employed in 3D Gaussian compression~\cite{papantonakis2024reducing, lee2024compactrad, wang2024end2end, fan2024lightgaussian, navaneet2024compgs}, where a learnable codebook is used to quantize high-dimensional feature representations, achieving efficient compression. 
Some other methods map 3D Gaussian primitives onto feature planes~\cite{morgenstern2024compact_selforg,lee2025compression} to compress with conventional video codecs. The feature planes require end-to-end optimization to obtain more compressible representations.
Further research focuses on generating more compact Gaussian models by meticulously designing 3D representations. Lu \textit{et al.} propose Scaffold-GS~\cite{lu2024scaffold}, clustering Gaussian primitives into anchor points with neural features.
More recent approaches utilize structural priors such as hash grids and context models to exploit the redundancies among 3DGS primitives, achieving about 100$\times$ compression compared to the vanilla 3DGS~\cite{chen2024hac,wang2024contextgs,chen2025hac++}.

However, the aforementioned methods require an optimization process for each specific scene. They also require dense multi-view images as input. Both characteristics limit their applicability. To address this, Chen \textit{et al.} propose FCGS~\cite{chen2024fcgc}, a general-purpose feedforward compression framework. By incorporating multiple priors into the Gaussian mixture modeling, FCGS achieves efficient compression for arbitrary 3D Gaussian data, even outperforming many scene-specific optimization methods. Nevertheless, the generality of FCGS limits its ability to exploit the unique characteristics of different types of 3D Gaussian data, leaving room for further improvement in compression efficiency.


\begin{figure*}[t]
\centering
\includegraphics[width=0.98\textwidth]{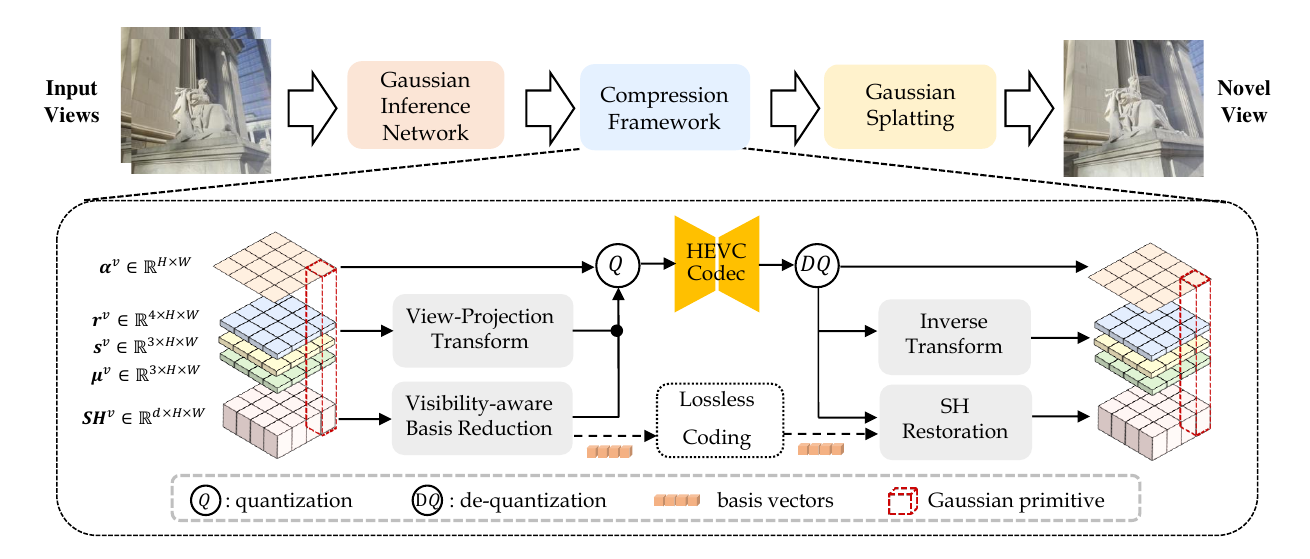}
\caption{The overall framework of TinySplat. We generate Gaussian feature maps from existing Gaussian inference networks in a feedforward manner and compress them with a meticulously designed compression framework. The symbols $\boldsymbol{\alpha}^v, \boldsymbol{r}^v, \boldsymbol{s}^v, \boldsymbol{\mu}^v, \boldsymbol{SH}^v$ denote the opacity, rotation, scale, mean position, and SH maps associated with view $v$, respectively. These parameters jointly define the geometry and appearance of the Gaussian primitives. In our compression framework, we first apply the VPT and VABR to reduce cross-channel redundancy. Subsequently, all features are quantized into 14-bit integers, and each feature plane is independently encoded as a grayscale image using the HEVC codec. On the decoder side, we perform dequantization and apply the inverse transforms to reconstruct the original Gaussian feature maps. Finally, we merge Gaussian maps from different views to form the complete 3D scene representation and render novel views.}
\label{fig:framework}
\end{figure*}

\begin{figure}[t]
\centering
\includegraphics[width=0.98\linewidth]{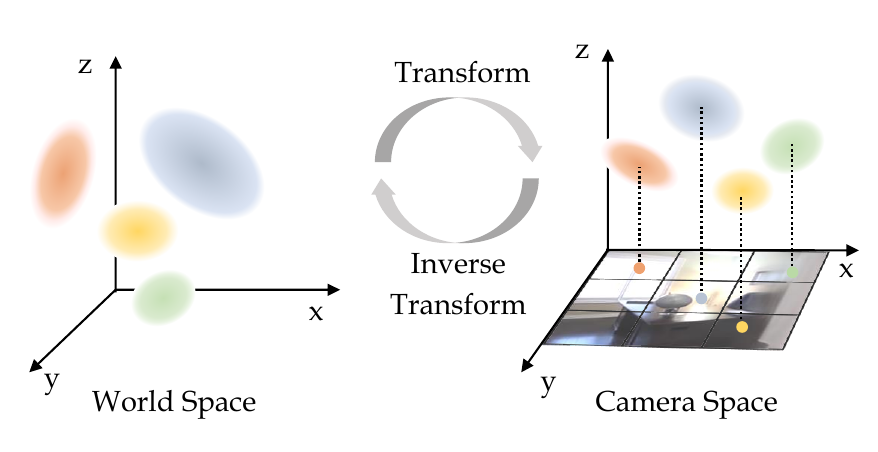}
\caption{Illustration of the proposed VPT. We project the feedforward-generated Gaussian primitives from world space into the corresponding input camera space. We perform compression in camera space, reduce inter-channel correlation of geometric parameters for better efficiency. During decoding, the geometry is transformed back into world space to reconstruct the 3D scene.
}
\label{fig:vp-trans}
\end{figure}

\section{Method}
\label{sec:method}

\subsection{Preliminary}
The widely recognized 3D Gaussian splatting employs a collection of Gaussian primitives to model 3D scenes. Each primitive consists of geometric and SH parameters that define its shape and radiance. Specifically, the geometric properties of Gaussian primitives are characterized by the Gaussian probability density function,
\begin{equation}
    \mathcal{G}(\boldsymbol{x}) = \frac{1}{(2\pi)^\frac{3}{2}|\boldsymbol{\Sigma}|^{\frac{1}{2}}}exp(-\frac{1}{2}(\boldsymbol{x}-\boldsymbol{\mu})^T\boldsymbol{\Sigma}^{-1}(\boldsymbol{x}-\boldsymbol{\mu})),
\end{equation}
where $\boldsymbol{\mu}$ and $\boldsymbol{x}$ denote the Cartesian coordinates of the Gaussian center and the sample point, respectively. $\boldsymbol{\Sigma}$ is the 3D covariance matrix and $|\boldsymbol{\Sigma}|$ denotes its determinant. In the vanilla 3DGS, the covariance matrix is further decomposed into rotation matrix $\boldsymbol{R}$ and scaling matrix $\boldsymbol{S}$,
\begin{equation}
    \boldsymbol{\Sigma} = \boldsymbol{RSS}^T\boldsymbol{R}^T,
\end{equation}
where the diagonal scaling matrix $\boldsymbol{S}$ is stored as a vector $\boldsymbol{s}\in \mathbb{R}^3$ and $\boldsymbol{R}$ is stored as quaternion $\boldsymbol{q}\in \mathbb{R}^4$ that represents rotation. 

In the rendering process, each Gaussian primitive is projected into the camera space and integrated to produce a 2D density function, which is combined with the opacity parameter $\sigma$ to compute pixel-wise opacity. Meanwhile, the color value is obtained by querying SH functions with the view direction.

To reconstruct a specific scene, the vanilla 3DGS approach first generates an initial set of Gaussian primitives via structure-from-motion techniques. Subsequently, this initial model undergoes optimization to match input ground-truth images, including adaptive densification guided by gradients to effectively capture fine geometric and textural details.

Despite delivering high-quality reconstructions, this optimization-based approach suffers from slow processing speeds. To mitigate this limitation, recent studies propose eliminating the computationally intensive gradient descent step by introducing deep NNs enhanced with cross-view attention mechanisms. These networks trained on extensive multi-view datasets can directly produce accurate 3D scene representations through feedforward inference. To facilitate NN processing, this approach typically generates Gaussian feature maps of the same resolution as the input images, where each location stores all geometric and color attributes of a single Gaussian. Formally, the Gaussian inference process can be expressed as,
\begin{equation}
\label{feedforward gaussian manner}
    \begin{aligned}
    \mathcal{F}: \{( \boldsymbol{I}^v, \boldsymbol{K}^v,& \boldsymbol{E}^v)\}_{v=1}^V \Rightarrow \\
    &\{\cup(\sigma_i^v, \boldsymbol{\mu}_i^v, \boldsymbol{q}_i^v, \boldsymbol{s}_i^v, \boldsymbol{SH}_i^v)\}_{i=1,...,H\times W}^{v=1,...,V},
\end{aligned}
\end{equation}
where $\boldsymbol{I}$ denotes the input image, $\boldsymbol{K}$, $\boldsymbol{E}$ denote the intrinsic and extrinsic parameters corresponding to the input image, respectively. $V$ denotes the number of viewpoints.

\subsection{Proposed Pipeline}
The overall pipeline of the proposed approach is illustrated in Fig.~\ref{fig:framework}. Initially, we employ pre-trained feedforward 3D Gaussian inference networks~\cite{mvsplat, xu2024depthsplat} to generate a set of 3D Gaussian primitives representing the target scene. For each input view, the network produces Gaussian primitives corresponding directly to image pixels, with each primitive consisting of geometric and color features as described in Eq.~\ref{feedforward gaussian manner}. These features collectively form per-view feature maps.

Compared to the original images, the size of this 3D Gaussian representation increases by more than two orders of magnitude. However, according to information theory principles, applying deterministic transformations does not inherently amplify the information content of the source data. Therefore, the generated 3D Gaussian model contains substantial redundancy, while existing compression methods for 3DGS relying on optimization are incompatible with the feedforward inference pipeline. To address this, we introduce a training-free compression framework designed to reduce both spatial and perceptual redundancies, producing a significantly more compact representation of the 3D scene.

Our compression framework comprises two lightweight yet highly effective modules, VPT and VABR. These modules specifically address the challenges posed by the absence of end-to-end optimization. First, high-dimensional Gaussian primitives inherently contain extensive redundancies. Second, without supervision from explicit distortion metrics, quantifying how deviations in parameters affect rendering quality becomes challenging.

The VPT targets geometric compression based on the following key observations. First, rendering occurs in camera space, making distortion measurements in this domain more directly representative of rendering quality impacts. Second, Gaussian positions derived via back-projection exhibit stronger spatial correlations in camera space. VPT leverages these properties by performing a reversible transformation of geometric features from world space into camera space, facilitating more efficient compression.

The VABR module addresses perceptual redundancy in appearance parameters by leveraging the anisotropic properties of SH functions. By assigning visibility-aware importance weights to SH coefficients and analyzing SH distributions across the whole scene, VABR derives perceptually consistent basis functions. This adaptive approach selectively retains color features that significantly contribute to dominant viewing directions, enabling efficient yet visually faithful compression.

Subsequently, we pre-quantize the transformed features into a 14-bit integer format based on their standard deviations. These quantized feature maps are then encoded as grayscale images using an off-the-shelf HEVC codec to further reduce spatial redundancy. Additionally, metadata required for reconstruction, including camera parameters for the inverse VPT, basis vectors for inverse VABR, and other dequantization-related information, is losslessly encoded and transmitted to the decoder.

On the decoder side, we first decode the feature maps using an HEVC decoder, followed by dequantization to recover the floating-point features. The reconstructed geometric parameters are then transformed back to the original world coordinate system through inverse VPT. Concurrently, the color feature coefficients are restored to their original SH representation using the received basis vectors. Finally, by integrating these decoded attributes, we reconstruct the complete 3D Gaussian representation, enabling the rendering of novel views using a standard 3DGS renderer.



\begin{figure}[t]
\centering
\includegraphics[width=0.9\linewidth]{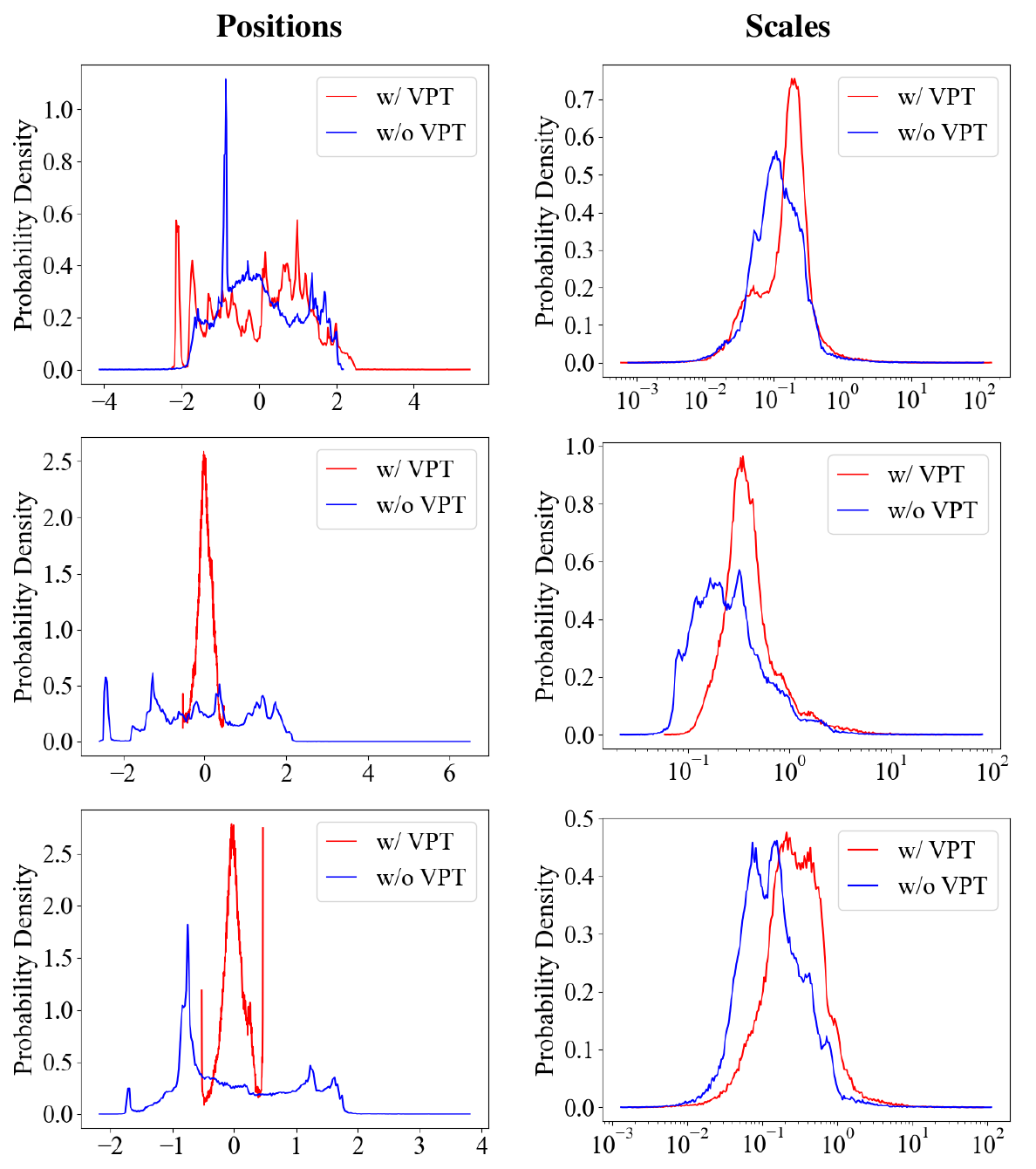}
\caption{Statistical distributions of positions and scaling factors before and after VPT. After applying VPT, most of the positional energy is concentrated in the depth channel~(top row), and the distribution of scaling factors becomes more compact.
}
\label{fig:distribution}
\end{figure}

\subsection{View-projection Transform}

\label{sec_proj_trans}
The Gaussian models generated by the inference network typically exhibit pixel-wise alignment with the input views, distinguishing them from conventional 3D Gaussian representations. This characteristic introduces a highly structured spatial arrangement into the generated models. Specifically, each Gaussian center is computed by back-projection, resulting in an ordered distribution of 3D positions. Moreover, the projected areas of these Gaussian ellipsoids remain relatively uniform across the input view, causing the scale parameters to correlate strongly with their distances to the camera. This reflects a systematic depth-dependent scaling pattern.


Based on the above analysis, we propose to apply the VPT to the geometry parameters of Gaussian primitives, conditioned upon input camera parameters.
A conceptual illustration of the VPT is shown in Fig.~\ref{fig:vp-trans}, where we transform the Gaussian geometric parameters into the corresponding camera space.
Specifically, we first apply the view transformation to the position of Gaussian centers,
\begin{equation}
    z_{i,j}^v(x_{i,j}^v,y_{i,j}^v,1)^T = \boldsymbol{K}^v\cdot(\boldsymbol{R}^v\cdot \boldsymbol{\mu}_{i,j}^v + \boldsymbol{T}^v),
\end{equation}
where $\boldsymbol{K}^v$ denotes the intrinsic matrix of the input view $v$, while $\boldsymbol{R}^v$ and $\boldsymbol{T}^v$ are the rotation matrix and translation vector derived from the extrinsic parameters.
The vector $\boldsymbol{\mu}_{i,j}^v$ represents the Gaussian center in world coordinates. $z_{i,j}^v$, $x_{i,j}^v$, and $y_{i,j}^v$ represent the transformed position. In particular, $z_{i,j}^v$ corresponds to the depth map, while $x_{i,j}^v$ and $y_{i,j}^v$ represent normalized 2D coordinates in the image plane.
Given the pixel-aligned nature of feedforward-generated Gaussians, coordinates $(x_{i,j}^{v}, y_{i,j}^{v})$ closely cluster around their corresponding pixel centers $(\frac{i}{H}, \frac{j}{W})$, with $H$ and $W$ being the height and width of the input images, respectively. Using this regularity, we encode only the offsets relative to pixel centers to enhance compression efficiency.

After transforming center positions into camera space, we approximate transformations for the Gaussian shape parameters. Since ellipsoids do not maintain exact geometric forms under perspective projection, we perform the following approximations,
\begin{equation}
    \hat{\boldsymbol{q}}_{i,j}^{v}=Quat(\boldsymbol{R}^v)\cdot \boldsymbol{q}_{i,j}^v,
\end{equation}
\begin{equation}
    \hat{\boldsymbol{s}}_{i,j}^{v}=f\frac{\boldsymbol{s}_{i,j}^{v}}{z_{i,j}^{v}},
\end{equation}
where $\hat{\boldsymbol{q}}_{i,j}^{v}$ and $\boldsymbol{q}_{i,j}^v$ denotes the rotation quaternions before and after the transformation, respectively, and $\hat{\boldsymbol{s}}_{i,j}^{v}$ and $\boldsymbol{s}_{i,j}^v$ denote the corresponding scaling factors. $Quat(\boldsymbol{R}^v)$ denotes the quaternion form of $\boldsymbol{R}^v$, $f$ represents the focal length.
The transformation of the rotation and scale parameters can be interpreted as rotating the Gaussian ellipsoids into the camera coordinate system, followed by a perspective scaling toward the focal plane.
Since all of these transformations are invertible, the approximations involved do not compromise rendering performance. 

To illustrate the effectiveness of VPT, we analyze the distribution of Gaussian geometric parameters before and after the transformation, as depicted in Fig.~\ref{fig:distribution}. Compared to the initial distribution, the transformed Gaussian geometric parameters exhibit enhanced regularity and spatial coherence.


\begin{figure}[t]
    \centering

    \subfloat{%
        \includegraphics[width=0.95\linewidth]{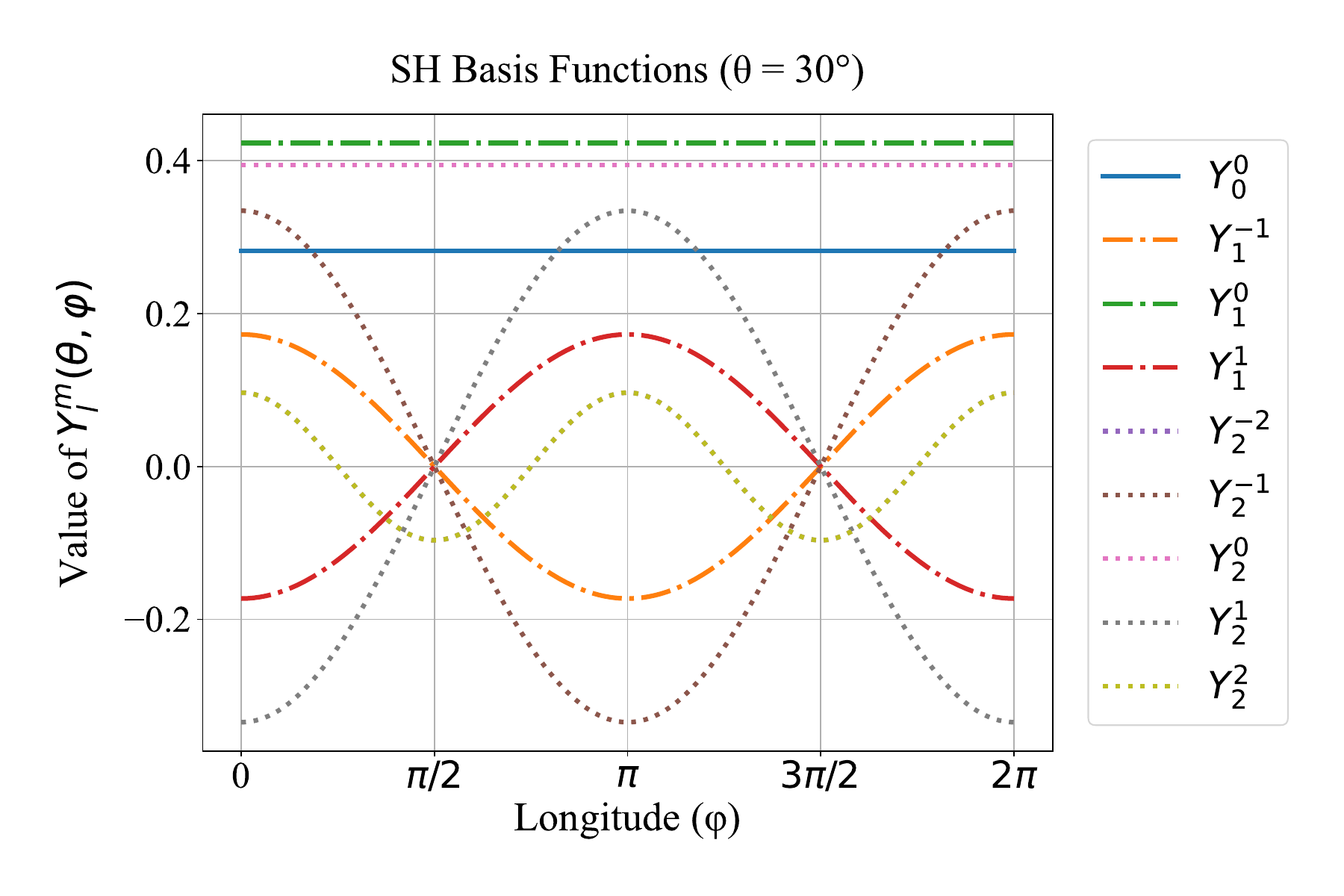}%
        \label{fig:sh_30}%
    }

    \vspace{-0.5cm}
    
    \subfloat{%
        \includegraphics[width=0.95\linewidth]{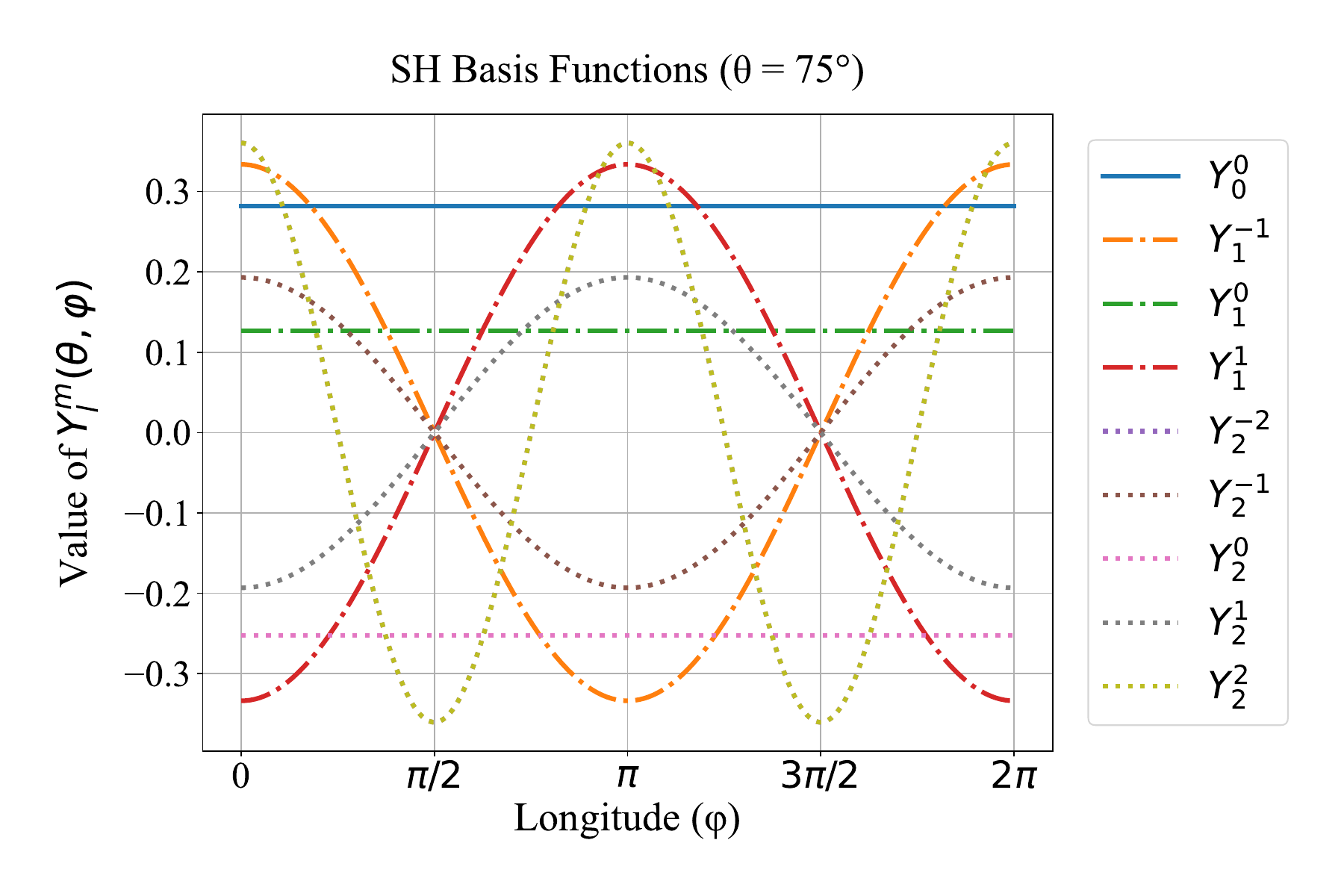}%
        \label{fig:sh_75}%
    }

    \vspace{-0.2cm}
    
    \caption{The variation of SH basis function values with respect to viewing direction. At a specific direction, a larger absolute value indicates that the corresponding SH coefficient has a greater influence on the final rendering result. $Y_l^m$ in the figure denotes the SH basis function of degree $l$, where $m$ ranges from $-l$ to $l$, representing $2l+1$ distinct spatial orientations.}
    \label{fig:sh_fig}
\end{figure}

\subsection{Visibility-aware Basis Reduction}
\label{sec_da_PCA}


3DGS employs SH feature coefficients to model the anisotropic radiance distribution of each Gaussian ellipsoid. Although SH functions can effectively model the radiance with strong physical interpretability, from a compression perspective, the representation is notably redundant. Representing RGB colors using SH functions of order $N_l$ requires a total of $3 \cdot (N_l + 1)^2$ coefficients, leading to substantial storage overhead. However, Many SH basis functions correspond to spherical regions rarely or never observed from the available viewpoints, thus minimally contributing to rendering quality.

The redundancy is particularly pronounced in sparse view reconstruction tasks, where the range of novel views is limited due to the lack of visual information from distant viewpoints. 
As illustrated in Fig.~\ref{fig:sh_fig}, the response of SH functions varies significantly across viewing angles. Therefore, treating all SH coefficients equally without considering their visibility may reduce the compression efficiency.

To address this issue, we introduce the VABR, which linearly adapts standard SH bases into a compact, visibility-aware basis set. We first compute the visibility-based weights to capture the directional importance of each SH component. Each weight factor $\lambda_l^m$ is defined as the average absolute value of the corresponding SH basis function $Y_l^m$ over a region of valid directions,
\begin{equation}
    \lambda_l^m=\mathbb{E}_{\boldsymbol{D}}|Y_l^m(\boldsymbol{\omega})|=\frac{1}{|\boldsymbol{D}|}\int_{\boldsymbol{\omega} \in \boldsymbol{D}}|Y_l^m (\boldsymbol{\omega})|d\boldsymbol{\omega},
\end{equation}
where $l$ and $m$ represent the order and orientation factors of the SH function. $\boldsymbol{D}$ denotes the area of valid directions. In practice, it is challenging to rigorously define the area $\boldsymbol{D}$, so we approximate it via Monte Carlo integration,
\begin{equation}
    \lambda_l^m=\frac{1}{N_s}\sum_{i=1}^{N_s}|Y_l^m(\boldsymbol{\omega}_i)|,
\end{equation}
where $N_{s}$ is the number of sample directions.
To obtain representative viewing directions, we cast camera rays through a $3 \times 3$ grid uniformly covering the entire image plane for each input view and aggregate them.

Subsequently, these weights inform the construction of a compact set of new spherical basis functions, which also reflect the statistical distribution of SH coefficients within the scene.
In particular, to ensure balanced perceptual contributions, we scale the original SH bases by their visibility-aware weights and linearly combine them into new optimized bases, resulting in the following transformation,
\begin{equation}
    \hat{\boldsymbol{Y}}(\boldsymbol{\omega}) = \boldsymbol{W}^T \cdot\boldsymbol{\Delta}^{-1}\cdot \boldsymbol{Y}(\boldsymbol{\omega}),
\end{equation}
where $\boldsymbol{Y}(\boldsymbol{\omega})$ and $\hat{\boldsymbol{Y}}(\boldsymbol{\omega})$ denote the original SH basis functions and the transformed basis functions, respectively. The weight matrix $\boldsymbol{\Delta}$ denotes the diagonal matrix consisting of $(\lambda_0^0, ...,\lambda_l^m,...)$. $\boldsymbol{W}\in \mathbb{R}^{d\times k}$ represents the transform matrix to be solved, which contains a set of d-dimensional unit orthogonal vectors, with $k$ representing the dimensionality of the transformed function space.
Subsequently, the forward and inverse transformations of the color coefficients are as follows,
\begin{equation}
    \boldsymbol{Z}=\boldsymbol{W}^{T}\cdot \boldsymbol{\Delta}\cdot  \boldsymbol{X},\ \ \hat{\boldsymbol{X}}=\boldsymbol{\Delta}^{-1}\cdot \boldsymbol{W}\cdot  \boldsymbol{Z},
\end{equation}
where $\boldsymbol{X},\hat{\boldsymbol{X}}\in \mathbb{R}^{d\times M}, \boldsymbol{Z}\in \mathbb{R}^{k\times M}$ represent the input, reconstructed, and transformed data matrices, respectively. Here, $M$ refers to the total number of Gaussian ellipsoids in the current scene.

To determine the transformation matrix $\boldsymbol{W}$ that can optimally preserve the perceptual information from higher-dimensional SH coefficients, we perform a statistical analysis on all SH coefficients in the current scene. Motivated by the formulation of principal component analysis, we compute the covariance matrix of $\boldsymbol{\Delta}\cdot \boldsymbol{X}$ and extract its principal components via eigen decomposition.
In particular, the eigenvectors corresponding to the $k$ largest eigenvalues are selected as the principal basis to form the transform matrix $\boldsymbol{W}$.






\begin{table*}[t]
  \renewcommand{\arraystretch}{1.3}
  \centering
  \caption{Quantitative comparison with 2 input views. TinySplat is compared with uncompressed models and FCGS on 3D Gaussians generated by different feedforward methods. ``Raw'' indicates the uncompressed baseline.}
    \begin{tabular}{cccc cc cc cc}
    \toprule
    \multirow{2}[1]{*}{\shortstack{Inference}} & \multirow{2}[1]{*}{\shortstack{Compression}} & \multicolumn{4}{c}{Re10k}     & \multicolumn{4}{c}{ACID} \\
\cmidrule{3-10}    & & PSNR(dB)$\uparrow$  & SSIM$\uparrow$  & LPIPS$\downarrow$ & Size(MB)$\downarrow$ & PSNR(dB)$\uparrow$  & SSIM$\uparrow$  & LPIPS$\downarrow$ & Size(MB)$\downarrow$ \\
    \midrule
    \multirow{2}[1]{*}{MVSplat} & Raw     & 26.36  & 0.8679  & 0.1291  & ~43   & 28.21  & 0.8419  & 0.1448  & ~43 \\
          & FCGS  & 25.90  & 0.8586  & 0.1533  & 2.650  & 27.61  & 0.8298  & 0.1800  & 2.596  \\
    (CVPR 2024) & Ours  & 26.32  & 0.8671  & 0.1349  & 0.216  & 28.13  & 0.8401  & 0.1528  & 0.229  \\
    \midrule
    \multirow{2}[1]{*}{DepthSplat} & Raw     & 27.50  & 0.8900  & 0.1125  & ~19   & 28.37  & 0.8481  & 0.1417  & ~19 \\
          & FCGS  & 27.35  & 0.8863  & 0.1201  & 2.674  & 28.27  & 0.8454  & 0.1466  & 2.666  \\
    (CVPR 2025) & Ours  & 27.43  & 0.8886  & 0.1192  & 0.166  & 28.29  & 0.8470  & 0.1501  & 0.181  \\
    \bottomrule
    \end{tabular}%
  \label{tab:main performance}%
\end{table*}%

\begin{table}[t]
  \centering
  \caption{$\alpha$ and $Q_{c}$ configurations for all data channels.}
    \begin{tabular}{ccccccc}
    \toprule
          & depth & offset xy & scale & rotation & color & opacity \\
    \midrule
    $N_{dim}$ & 1     & 2     & 3     & 4     & 6     & 1 \\
    \midrule
    $\alpha$ & 2048  & 256   & 256   & 256   & 1024  & 256 \\
    \midrule
    $Q_{c}$ & -4    & 12    & 0     & 9     & 3     & 0 \\
    \bottomrule
    \end{tabular}%
  \label{tab:qstep cfg}%
\end{table}%

\section{Experimental Results}
\label{sec:experiment}
\textbf{Datasets.} 
We conduct comprehensive experiments on several widely used benchmark datasets in the field. Among them, RealEstate10K~\cite{zhou2018stereo} and ACID~\cite{liu2021infinite} represent two large-scale multi-view 3D scene datasets that have been extensively adopted in related research. RealEstate10K dataset primarily comprises indoor scene multi-view images collected from online sources, containing 7,289 test scenes. The ACID dataset consists of numerous outdoor scenes captured by aerial drones, with 1,972 scenes for testing. In addition, we evaluate our method on a subset of the DL3DV dataset, containing 140 test scenes. Our test condition strictly adheres to the common configurations adopted in previous studies~\cite{pixelsplat, mvsplat, xu2024depthsplat}.


\textbf{Baseline.}
In the Gaussian generation stage, we adopt representative feedforward inference models, including MVSplat~\cite{mvsplat} and DepthSplat~\cite{xu2024depthsplat}, to generate 3D Gaussian feature maps. To demonstrate the generality and efficiency of the proposed pipeline, we compare our method with the optimization-free compression approach FCGS~\cite{chen2024fcgc} in terms of compression efficiency.

\textbf{Implementation details.}
In our experiments, the reserved color feature dimension $k$ for VABR is fixed at 6.
For 14-bit quantization, the quantization step for each channel is configured as $\sigma/\alpha$, where $\sigma$ represents the standard deviation of the current channel and $\alpha$ is an empirically chosen scaling factor, as detailed in Table~\ref{tab:qstep cfg}. 
We initially set $\alpha$ to 256 for each channel, and selectively increase it for a few error-sensitive channels to improve reconstruction quality.
Due to the varied distributions across different color basis components, we adopt a shared quantization step for all color channels to ensure consistent quantization. Here, $\sigma$ represents the standard deviation of the most dominant component.
After quantization, an offset is applied so that the minimum quantized value becomes $0$.
Extreme outliers in each channel are truncated to ensure that the maximum quantized values do not exceed the valid 14-bit range.

In the video coding module, each channel is treated as an independent grayscale image and encoded using the HEVC reference software HTM15.0-RExt8.1, with all channels encoded in parallel via separate processes. To achieve different compression rates, we adjust a global Quantization Parameter~(QP), denoted as $Q_g$. Then, a set of per-channel QP offsets $Q_{c}$ is experimentally configured, as summarized in Table~\ref{tab:qstep cfg}. Subsequently, the QP values for the HEVC codec are configured as $Q_c+Q_g$.
All experiments are conducted on a workstation equipped with an Intel Core i7-13700K CPU and a single NVIDIA RTX 4090 graphics card, running in a WSL2.0 environment with Ubuntu 20.04.

The overall syntax elements include Gaussian geometric parameters (position, scale, and rotation), color feature coefficients, and opacity.
In addition, a small amount of metadata is explicitly signaled to the decoder, including camera parameters, the basis vectors used for dimensionality reduction, as well as per-channel quantization steps and offset values.


\begin{figure*}[t]
\centering
\includegraphics[width=0.98\textwidth]{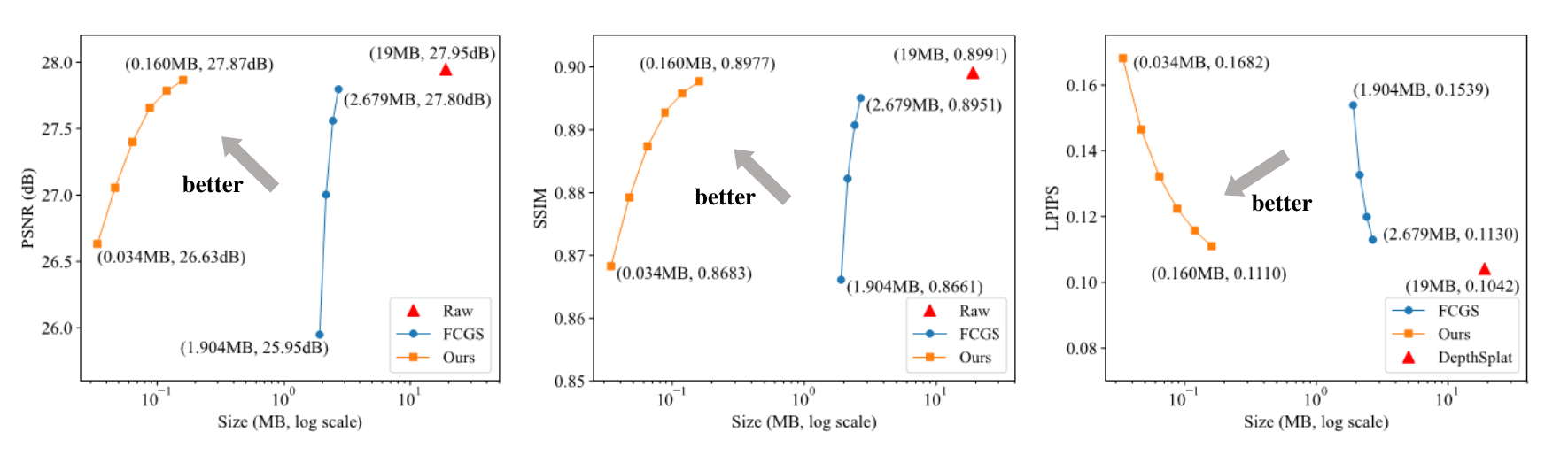}
\caption{Rate-distortion performance on the Re10K dataset, where distortion is measured in terms of novel view synthesis quality. "Raw" denotes the uncompressed 3D Gaussian model from DepthSplat. Compared to the SoTA method, our method achieves comparable quality (PSNR, SSIM, and LPIPS) with only 6\% of the storage size.}
\label{fig:rd_vs_FCGS}
\end{figure*}

\begin{table}[t]
  \centering
  \caption{Compression performance under different numbers of input views. ``Raw'' indicates the uncompressed baseline generated via Depthsplat.}
  \vspace{0.5em}
  \footnotesize
  \begin{tabular}{cccccc
                }
    \toprule
    Method & Views & PSNR (dB)$\uparrow$ & SSIM$\uparrow$ & LPIPS$\downarrow$ & Size (MB)$\downarrow$ \\
    \midrule
    Raw   & \multirow{3}[2]{*}{2} & 19.62  & 0.6265  & 0.2986  & ~33 \\
    FCGS  &       & 19.26  & 0.6183  & 0.3061  & 4.518  \\
    Ours  &       & 19.52  & 0.6196  & 0.3147  & 0.615  \\
    \midrule
    Raw   & \multirow{3}[2]{*}{4} & 23.16  & 0.7792  & 0.1745  & ~66 \\
    FCGS  &       & 22.23  & 0.7619  & 0.1897  & 9.054  \\
    Ours  &       & 22.99  & 0.7694  & 0.1923  & 1.275  \\
    \midrule
    Raw   & \multirow{3}[2]{*}{6} & 24.16  & 0.8191  & 0.1456  & ~100 \\
    FCGS  &       & 22.74  & 0.7943  & 0.1672  & 13.561  \\
    Ours  &       & 23.95  & 0.8085  & 0.1633  & 1.892  \\
    \bottomrule
  \end{tabular}
  \label{tab:compression_views}
\end{table}

\subsection{Objective Results}

\textbf{Overall Compression Performance.}
We first generate the 3D Gaussian models using 
the official implementations and pre-trained weights MVSplat\footnote{\url{https://github.com/donydchen/mvsplat}}~(SHA: 1f5e5486) and DepthSplat\footnote{\url{https://github.com/cvg/depthsplat}}~(SHA: 175b17a6). Then we compress the generated models using our compression framework to evaluate its effectiveness. 

We evaluate the rendering PSNR, SSIM, LPIPS, and compressed model size on two standard datasets with 2 input views and 3 target views, as summarized in Table~\ref{tab:main performance}. We employ a near-lossless compression configuration in our TinySplat, where $Q_g$ is set to 0. For the SoTA method FCGS, we faithfully reproduce its performance using the official implementation\footnote{\url{https://github.com/YihangChen-ee/FCGS}} to ensure a fair comparison. 
The results show that TinySplat achieves over 100$\times$ compression on the 3D Gaussian data produced by the baseline, with negligible loss in rendering quality. Compared to the SoTA training-free method, our approach still achieves a 15$\times$ higher compression ratio, while providing comparable rendering quality. 
The rendering quality suffers a noticeable drop when using the FCGS pipeline to compress Gaussian models from MVSplat. This performance degradation could be attributed to the distinctive distribution of 3D Gaussian data generated by MVSplat, as illustrated in the FCGS paper~\cite{chen2024fcgc}.
We highlight that we evaluate the performance on the complete datasets following the general configuration and report the average results to ensure a comprehensive and reliable comparison.

Beyond the two-view testing configuration, we also evaluate the compression performance under varying numbers of input views on the DL3DV dataset, as shown in Table~\ref{tab:compression_views}. We follow the testing configuration in DepthSplat~\cite{xu2024depthsplat} and reproduce the reported performance. The results demonstrate that our TinySplat generalizes well, as no significant degradation in compression quality is observed with more input views.

\textbf{Rate-Distortion Performance.}
To evaluate the impact of compression distortion on the rendering quality of 3D Gaussian models, we compare the rate-distortion performance of our proposed method against FCGS, as shown in Fig.~\ref{fig:rd_vs_FCGS}. To reduce the evaluation cost, we only conduct experiments on the first 200 test scenes from the Re10K and ACID datasets. The results show that our method achieves comparable or even better PSNR, SSIM, and LPIPS performance compared to FCGS with only 6\% of its storage space.

\begin{table}[t]
  \centering
  \caption{Runtime analysis. Our TinySplat is decomposed into key components. }
    \begin{tabular}{cccccc}
    \toprule
    \multirow{2}[4]{*}{Method} & \multicolumn{4}{c}{Encoding Time~(s)} & Decoding  \\
\cmidrule{2-5}          & VPT  & VABR & HEVC  & Total & Time (s) \\
    \midrule
    FCGS  & - & - & - & 4.074    & 4.095 \\
    TinySplat  & 0.003 & 0.004 & 0.971 & \textbf{1.016} & \textbf{0.042} \\
    \bottomrule
    \end{tabular}%
  \label{tab:time}%
\end{table}%

\textbf{Runing time Analysis.}
To assess the computational cost of our TinySplat, we measure the average encoding and decoding times, as reported in Table~\ref{tab:time}. TinySplat encodes a scene in approximately 1 second and supports real-time decoding. In contrast, the SoTA approach, which heavily relies on NN–based probabilistic models, requires about 4$\times$ more time for encoding and over 100$\times$ more for decoding.

The primary computational overhead in our pipeline arises from the HEVC video encoder. However, we note that the HEVC codec in our implementation is the reference software, which prioritizes coding efficiency and correctness over running speed. In practice, significantly faster encoding and decoding can be achieved through engineering optimizations~\cite{heng2014highly} or dedicated hardware implementations~\cite{7100895, 6317152}.

\begin{table}[t]
  \centering
  \caption{Bit allocation under different compression ratios.}
    \begin{tabular}{ccccccc}
    \toprule
    $Q_g$    & Size~(KB) & position & scale & rotation & color & opacity \\
    \midrule
    0     & 164.63  & 21.4\% & 17.5\% & 13.0\% & 34.6\% & 13.5\% \\
    3     & 122.29  & 21.0\% & 18.0\% & 11.5\% & 35.7\% & 13.7\% \\
    6     & 89.91  & 20.5\% & 18.8\% & 10.2\% & 36.7\% & 13.8\% \\
    12    & 47.90  & 20.0\% & 20.1\% & 8.4\% & 39.0\% & 12.6\% \\
    \bottomrule
    \end{tabular}%
  \label{tab:bit_allocation}%
\end{table}%

\begin{figure*}[t]
\centering
\includegraphics[width=\linewidth]{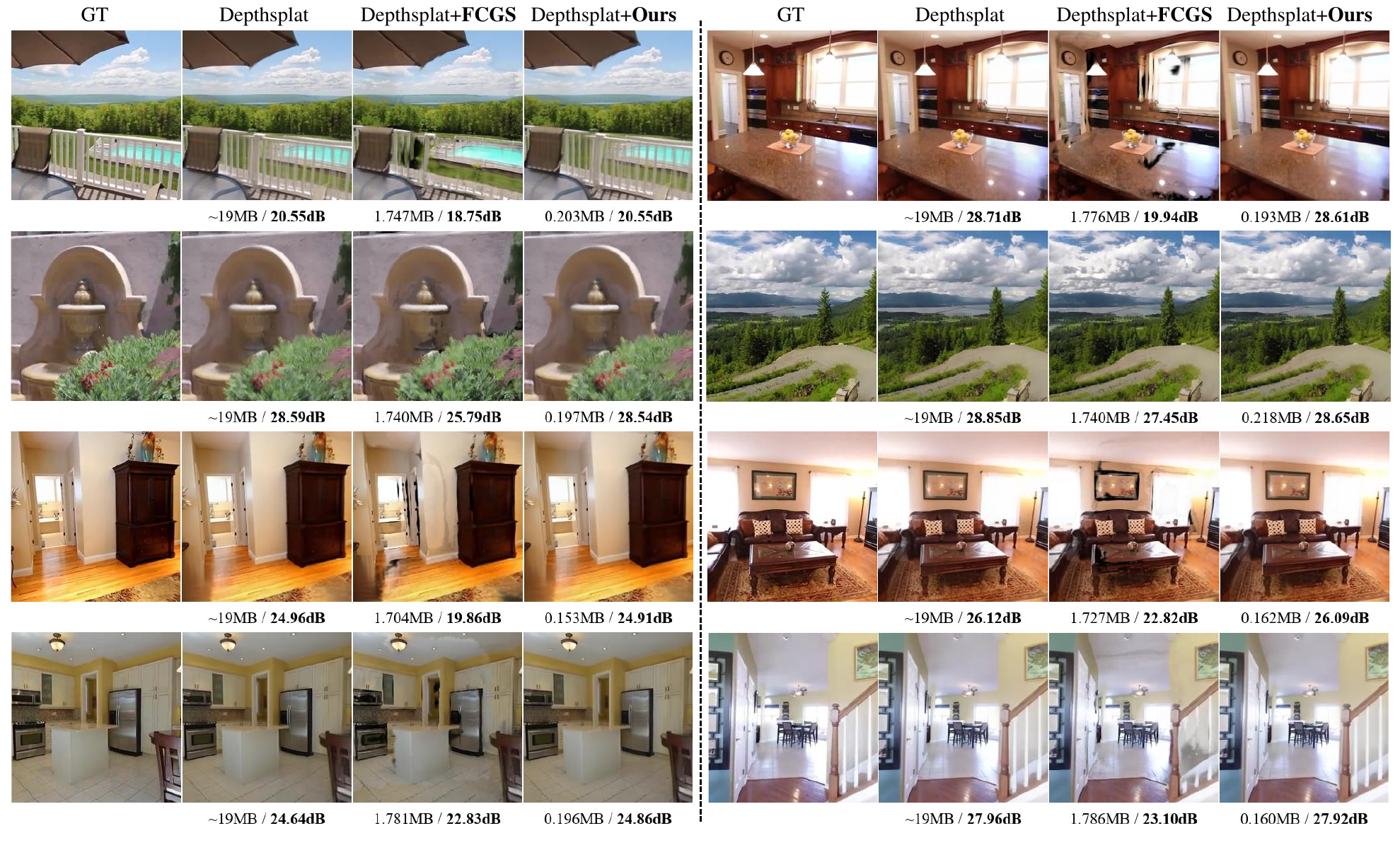}
\caption{Subjective results on Re10k dataset. We compress the 3D Gaussian models from DepthSplat with both FCGS and the proposed TinySplat. FCGS exhibits severe watermark-like artifacts at 11$\times$ compression ratio.
In contrast, our method achieves up to 100$\times$ compression while maintaining visually indistinguishable rendering quality from the uncompressed reference.}
\label{fig:subjective}
\end{figure*}

\begin{figure}[t]
\centering
\includegraphics[width=0.9\linewidth]{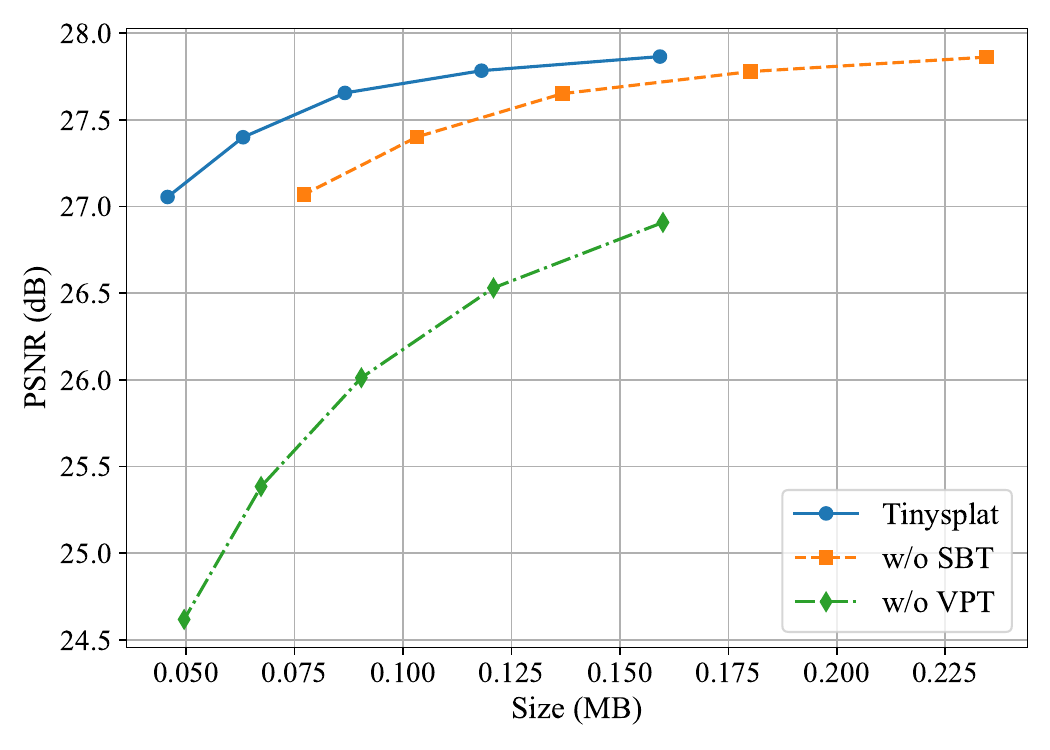}
\caption{Component-wise ablation analysis. We separately disable the VPT module and the VABR module to test their impact on coding efficiency. The PSNR is measured between the rendered novel view and the ground truth.}
\label{fig:ablation}
\end{figure}

\begin{figure*}[t]
\centering
\includegraphics[width=0.98\linewidth]{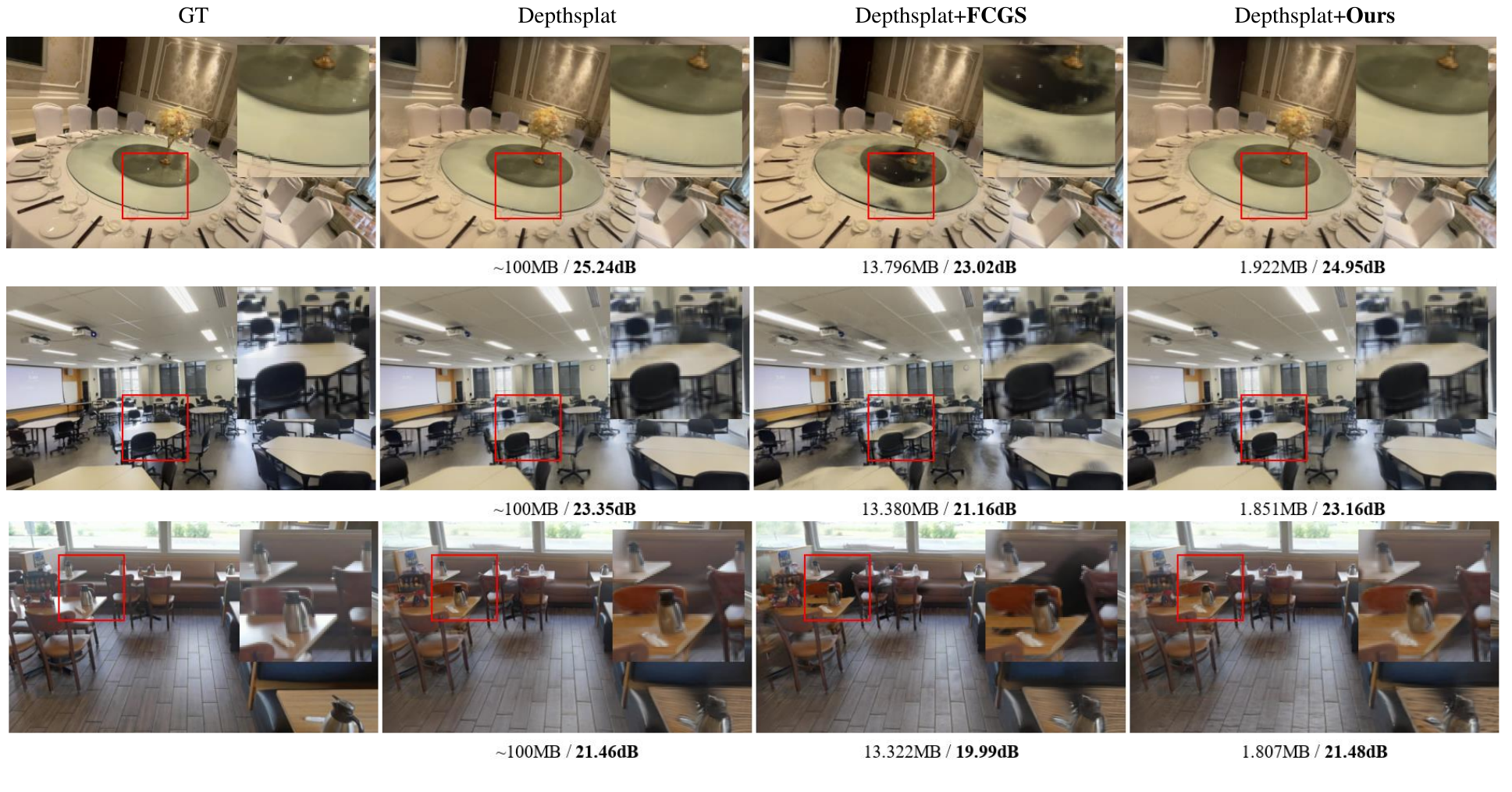}
\caption{Subjective results under 6 input views on DL3DV dataset. Our TinySplat achieves superior objective and perceptual quality using only 15\% of the storage compared to the SoTA compression method. FCGS exhibits noticeable compression artifacts even at relatively high bitrates.
}
\label{fig:subjective-6v}
\end{figure*}

\textbf{Bit Allocation.}
To better understand the bit allocation characteristics of our framework and guide future optimizations, we encode 200 scenes under various QP settings and analyze the average storage distribution across different components, as summarized in Table~\ref{tab:bit_allocation}. The results show that the relative bit allocation remains largely consistent across compression ratios, with color parameters consistently occupying the largest share of the bitrate.

\subsection{Subjective Results}
We adopt DepthSplat as the representative Gaussian inference method and compare subjective compression artifacts using FCGS and our proposed framework, as shown in Fig.~\ref{fig:subjective}.
In particular, we adopt the highest compression ratio setting offered by FCGS, which still results in a storage size several times larger than that of TinySplat.
The rendering results show that FCGS easily introduces prominent watermark-like artifacts, whereas TinySplat maintains visually negligible degradation, even at higher compression ratios.

We also evaluate the subjective performance under 6 input views, as illustrated in Fig.~\ref{fig:subjective-6v}. The results demonstrate that increasing the number of input views does not noticeably degrade the perceptual quality of our method, highlighting its strong generalization capability. In contrast, FCGS produces severe artifacts even at relatively low compression rates.

\subsection{Ablation Studies}
\textbf{Component-wise ablation analysis.} We conduct an ablation study by separately disabling the proposed techniques to assess their impact on the compression performance, as shown in Fig.~\ref{fig:ablation}. We utilize 200 scenes from the Re10k dataset for testing. Results demonstrate that VPT can effectively reduce the loss of geometric information under the same bitrate, thereby improving rendering quality.
Meanwhile, VABR enhances compression efficiency by concentrating the energy of SH coefficients, enabling significantly higher compression ratios.



\begin{figure}[t]
\centering
\includegraphics[width=0.98\linewidth]{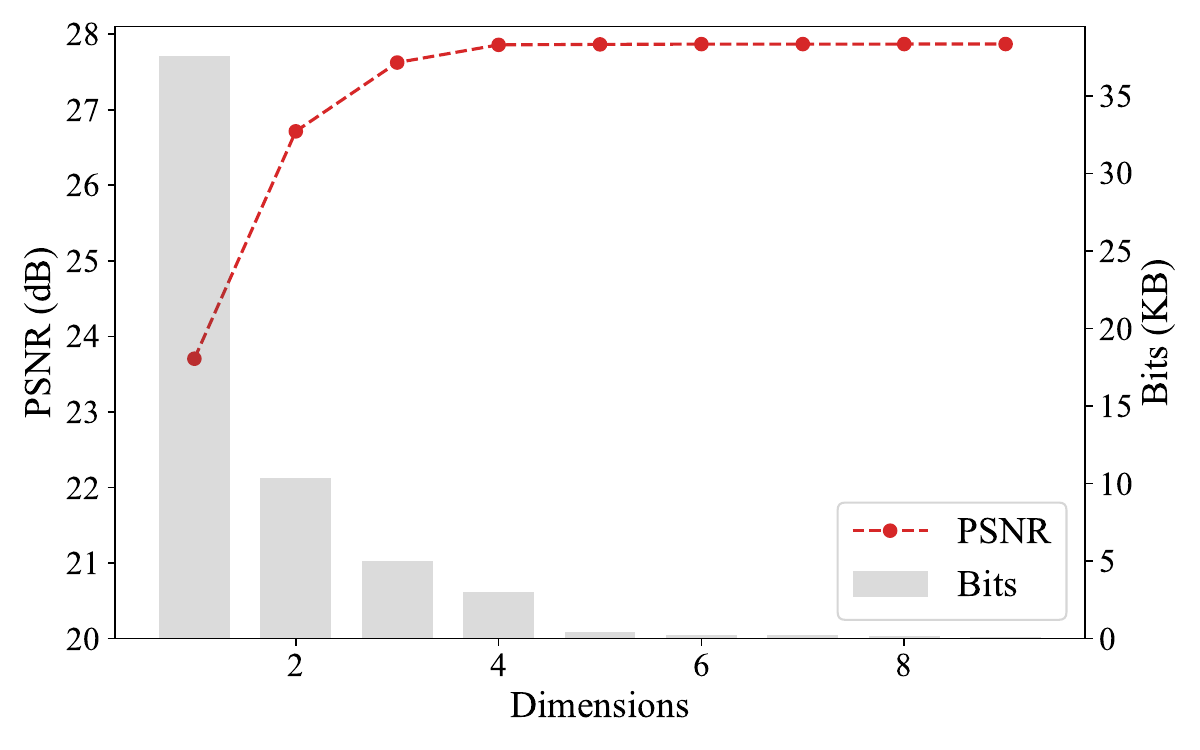}
\caption{Ablation study of the VABR. We present the variation of rendering PSNR with feature dimensionality increasing, along with the storage cost of each individual feature channel.
The results show that our method more effectively concentrates energy into the leading dimensions, enabling a more compact and efficient feature representation.}
\label{fig:SBR_ablation}
\end{figure}

\textbf{Ablation on VABR.} 
We conducted an experimental analysis to investigate the impact of the retained feature dimensions on compression efficiency. Starting from the most dominant component, we progressively added additional components to evaluate how encoding performance evolves.
In addition, we computed the storage cost associated with each individual dimension. As shown in Fig.~\ref{fig:SBR_ablation}, the results indicate that after applying the VABR, the first feature dimension concentrates the majority of the information, accounting for the largest portion of the total bitrate. The information content in subsequent dimensions decreases gradually. The overall rendering performance remains nearly unchanged as the feature dimensionality exceeds 6. These results validate the effectiveness of our VABR, which concentrates most of the meaningful information in the first few dimensions.


\section{Limitations}
\label{sec:limitations}
In this work, we do not modify existing 3D Gaussian inference networks; instead, we focus on developing a general compression framework applicable to the Gaussian data they generated. While our method effectively produces compact 3D Gaussian representations, there remains a notable gap between our achieved compression ratios and theoretical limits. Since 3D Gaussians are derived entirely from a few input images, information theory suggests that their entropy should not surpass that of the inputs. However, even after compression, the size of the 3D Gaussian models still exceeds that of the original images compressed with conventional codecs.
We attribute this discrepancy primarily to the inference process itself, where complex NN models introduce significant prior knowledge into the generated 3D Gaussian representations. Developing lightweight methods to explicitly extract and leverage these priors presents a promising direction for future research. In addition, redesigning or fine-tuning inference networks to generate inherently more compressible representations represents another important avenue for exploration.



\section{Conclusion}
\label{sec:conclusion}
In this paper, we presented TinySplat, a fully feedforward approach for generating compact 3D Gaussian scene representations directly from multi-view images without per-scene optimization. Bridging the gap between fast 3D Gaussian generation and practical deployment, TinySplat integrates a feedforward inference stage with a novel rendering-aware compression stage.
For the compression stage, we proposed a dedicated framework with VPT and VABR modules to reduce structural and perceptual redundancy, respectively. Furthermore, we employed a standard video codec to eliminate spatial redundancy. Extensive experiments demonstrated that TinySplat significantly reduces storage requirements while maintaining high rendering quality. 

These findings offer an intuitive understanding of the redundancy characteristics in feedforward-generated 3D Gaussian data, providing practical guidance for future exploration and system design.
We believe TinySplat paves the way for scalable and deployable 3D scene representation, and sets a solid foundation for future research in real-time neural graphics systems.

\bibliographystyle{IEEEtran}
\bibliography{TCSVT}

\vfill

\end{document}